\documentclass[10pt, a4paper]{article}

\usepackage[final]{lrec-coling2024}

\usepackage{amssymb}
\usepackage{amsmath}
\usepackage{algorithm,algorithmic}
\usepackage{bm}
\usepackage{booktabs}
\usepackage{overpic}
\usepackage{appendix}
\usepackage{graphicx}
\usepackage{tabularx}
\usepackage{soul}
\usepackage{ulem}

\usepackage{color}
\usepackage{xcolor}
\usepackage{xstring}
\usepackage{makecell}
\usepackage{hyperref}
 \definecolor{darkblue}{rgb}{0, 0, 0.5}
  \hypersetup{colorlinks=true, citecolor=darkblue, linkcolor=darkblue, urlcolor=darkblue}

\title{Making Pre-trained Language Models Better \\ Continual Few-Shot Relation Extractors}

\name{Shengkun Ma, Jiale Han, Yi Liang, Bo Cheng} 

\address{State Key Laboratory of Networking and Switching Technology, \\
         Beijing University of Posts and Telecommunications \\
         \{mashengkun, hanjl, liangyi, chengbo\}@bupt.edu.cn\\}

\abstract{
Continual Few-shot Relation Extraction (CFRE) is a practical problem that requires the model to continuously learn novel relations while avoiding forgetting old ones with few labeled training data. The primary challenges are catastrophic forgetting and overfitting. This paper harnesses prompt learning to explore the implicit capabilities of pre-trained language models to address the above two challenges, thereby making language models better continual few-shot relation extractors. Specifically, we propose a Contrastive Prompt Learning framework, which designs prompt representation to acquire more generalized knowledge that can be easily adapted to old and new categories, and margin-based contrastive learning to focus more on hard samples, therefore alleviating catastrophic forgetting and overfitting issues. To further remedy overfitting in low-resource scenarios, we introduce an effective memory augmentation strategy that employs well-crafted prompts to guide ChatGPT in generating diverse samples. Extensive experiments demonstrate that our method outperforms state-of-the-art methods by a large margin and significantly mitigates catastrophic forgetting and overfitting in low-resource scenarios.
\\ \newline 
}

\begin{document}

\maketitleabstract

\section{Introduction}
Relation Extraction (RE) is a fundamental and important task in the field of natural language processing, which aims to extract the underlying relation between entities in a sentence or document. Traditional RE methods \cite{peng2020learning, chen2022knowprompt} train models on a large number of labeled samples and subsequently test them on data with the same label space. However, in real-life scenarios where new relations emerge all the time, these models may experience a substantial performance decline when adapting to novel relations. In addition, these models heavily rely on the massive labeled data, which demands considerable time and effort to collect.

\begin{figure}[t]
  \flushleft
  \setlength{\abovecaptionskip}{-10pt}
  \includegraphics[width=0.5\textwidth]{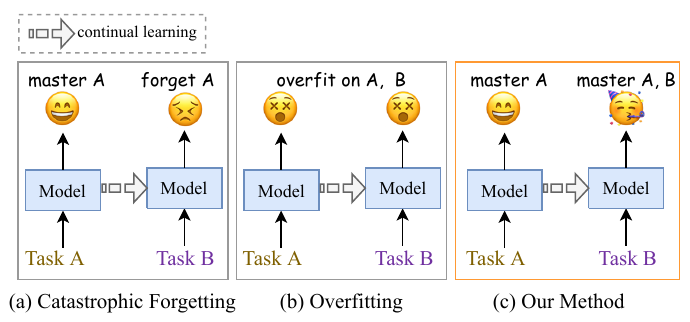}
  \caption{Overview of catastrophic forgetting and overfitting in CFRE.}
  \label{teaser}
\end{figure}

Therefore, Continual Few-Shot Relation Extraction (CFRE) \cite{qin2022continual} has been proposed, which aims to continually learn new relations while retaining knowledge of previously learned relations, all within the constraints of limited labeled data. This practical task brings forth two significant challenges: catastrophic forgetting and overfitting, as shown in Figure~\ref{teaser}.   Catastrophic forgetting refers to the phenomenon where the model abruptly forgets the knowledge gained from previous tasks when learning new tasks. It's worth noting that some latest studies \cite{luo2023empirical, zhai2023investigating} point out that catastrophic forgetting even exists in large language models, which makes this issue well worth studying. 
Overfitting occurs when a model learns to perform exceptionally well on the training data but fails to generalize effectively to unseen data due to fitting noise or irrelevant patterns, which tends to be more pronounced in low-resource scenarios with sparse training data.

There are several methods have been proposed to address CFRE. \citet{qin2022continual} tackle both issues by introducing embedding space regularization and data augmentation from external data sources. \citet{wang2023serial} introduce serial contrastive knowledge distillation to preserve the prior knowledge, thus addressing catastrophic forgetting. \citet{chen2023consistent} design a consistent prototype learning method to mitigate catastrophic forgetting. While these methods have demonstrated remarkable performance, they do not fully explore the rich inherent knowledge embedded within Pre-trained Language Models (PLMs) to effectively combat catastrophic forgetting and overfitting.

Recently, prompt learning is popular for its magic ability to unlock the potential of PLMs \cite{lester2021power} and has been widely proven to be simple and effective, especially in the few-shot setting \cite{gao2021making, gu2022ppt}.
In this work, we delve into the implicit capabilities of PLMs by leveraging prompt learning to address catastrophic forgetting and overfitting, thereby making language models better continuous few-shot relation extractors. To the best of our knowledge, this is the first exploration of prompt technologies in the CFRE task. Specifically, we propose a \textbf{C}ontrastive \textbf{P}rompt \textbf{L}earning framework (CPL), comprising a prompt representation module and a margin-based contrastive learning module. 
Prompt Representation introduces an effective semi-automated template for RE, which converts RE from classification to text infilling based on context. In this way, PLM tends to learn general task knowledge rather than specific relation categories, becoming adept at identifying previous and novel relations, thereby mitigating catastrophic forgetting.
Margin-based contrastive learning makes the model focus more on hard samples and thereby takes a more uniform feature distribution, which effectively alleviates overfitting.
To further remedy overfitting in low-resource scenarios, we leverage the power of large language models to bolster smaller PLMs and introduce an effective memory augmentation strategy, which employs well-crafted prompts to guide ChatGPT \cite{openai2022chatgpt} in generating diverse samples. During relation prediction phase, Nearest-Class-Mean classiﬁer is adopted instead of softmax classifier, which is more appropriate for incremental-class classification tasks. 

Extensive experiments on two popular RE datasets demonstrate our proposed approach significantly outperforms the baselines, for example, improving accuracy by 6.28\% compared to the state-of-the-art method in TACRED dataset. We also conduct a series of ablation studies to prove the effectiveness of each module. We release our code\footnote{\url{https://github.com/mashengkun/CPL}} to the community for future research. To summarize, our main contributions include:
\begin{itemize}
    \item We leverage prompt learning to explore the implicit capabilities of PLMs and propose CPL framework, combining it with a novel margin-based contrastive learning objective for CFRL, which alleviates catastrophic forgetting and overfitting issues simultaneously.
    \item We introduce a memory augmentation strategy by exploiting the power of LLMs to boost smaller PLMs, which employs well-crafted prompts to guide ChatGPT in generating samples and thus better combat overfitting. 
    \item Extensive experiments on two RE benchmarks show that our method outperforms SOTA models, proving the effectiveness of mitigating catastrophic forgetting and overfitting. 
\end{itemize}

\section{Related Work}
\subsection{Continual Learning}
Continual Learning (CL) aims to continually learn new knowledge from a sequence of tasks
while avoiding forgetting old knowledge.
The main challenge in CL is catastrophic forgetting \cite{mccloskey1989catastrophic}.
Existing CL methods are divided into three categories: 
1) regularization methods \cite{li2017learning, ritter2018online} use extra constraints to restrict the update of parameters, 
so that the models can remember more old knowledge. 
2) dynamic architecture methods \cite{fernando2017pathnet, mallya2018piggyback} extend model architecture dynamically to store new knowledge
when sequence tasks keep coming. 
3) memory-based methods \cite{rebuffi2017icarl, shin2017continual} store a few typical samples of current task to the memory,
and replay the memory after learning sequence task to review old knowledge.
Among these methods, memory-based methods are the most effective in NLP tasks \cite{wang2019sentence, han2020continual}.
However, data for new tasks is not always sufficient 
and also getting high-quality data tends to be expensive and time-consuming.
\citet{qin2022continual} first introduce Continual Relation Extraction (CRE) in the few-shot setting,
and \citet{wang2023serial, chen2023consistent} 
propose solutions based on memory methods. 
We also adopt memory-based strategy, 
but we concentrate more on how to better leverage PLMs to solve CFRE.

\subsection{Prompt Learning}

Prompt learning emerged with the birth of GPT-3 series \cite{brown2020language}
and has achieved remarkable performance in NLP tasks, especially in few-shot scenarios \cite{gao2021making, gu2022ppt}.
It reformulates downstream tasks into pre-training tasks by adding prompted tokens and guides PLMs to understand diverse tasks.
Previous prompt learning methods can be divided into three categories:
1) hard prompt \cite{schick2021exploiting} is to add handcrafted prompt tokens to the sentences and convert them into mask language modeling problem. Though effective, it requires sophisticated expert knowledge for different tasks which is cumbersome and time-consuming.
2) soft prompt \cite{lester2021power} is instead to add continuous trainable vectors contained in the sentences that can be automatically learned by the model.
However, models can not always learn the appropriate prompts without any prior expert knowledge, especially in low-resource scenarios.
3) hybrid prompt \cite{han2022ptr} combines untunable hard prompts and tuneable soft prompts, allowing the models to easily learn suitable templates with small manual intervention. It is verified to be the most effective prompt method according to recent studies \cite{gu2022ppt}. 
\citet{zhang2022prompt} introduce a prompt-based framework to enhance continual learning process in dealing with CRE. Different from them, we focus on the few-shot setting and adopt hybrid prompts to help with PLMs to alleviate catastrophic forgetting and overfitting.

\begin{figure*}[htbp]
  \centering
  \includegraphics[width=0.8\textwidth]{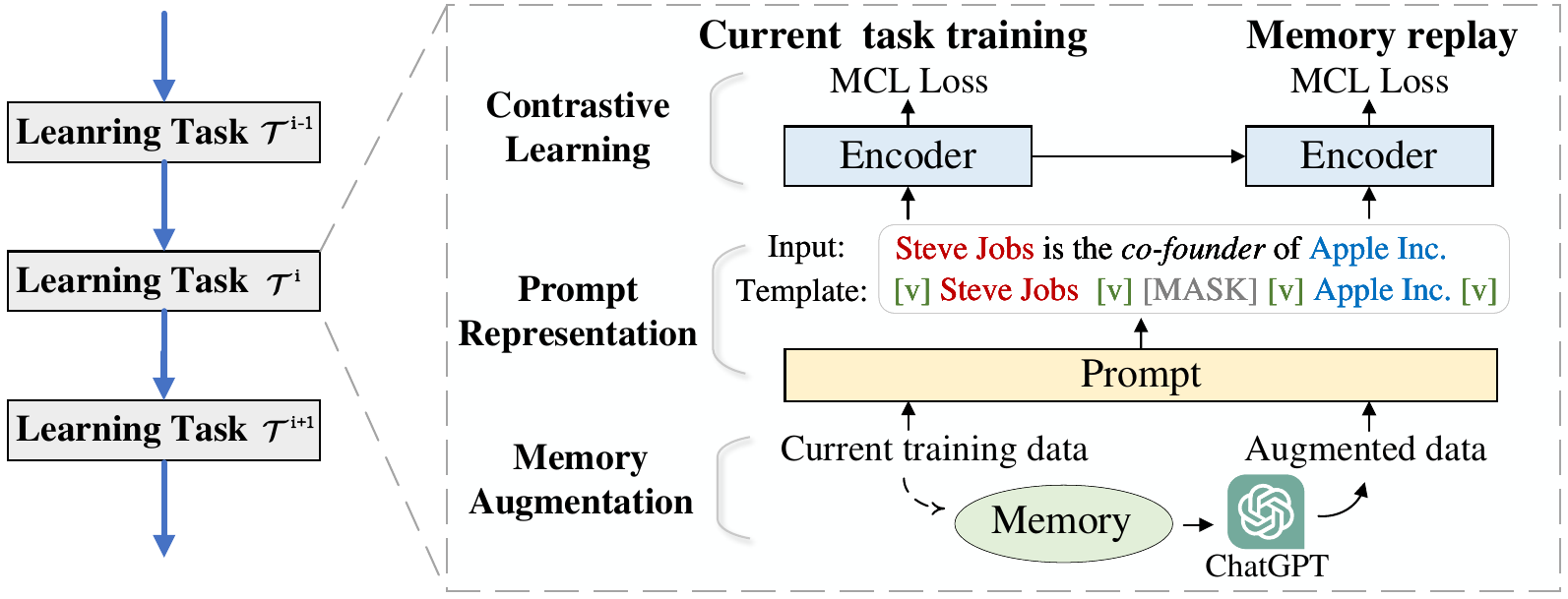}
  \caption{Framework of the proposed CPL.}
  \label{framework}
\end{figure*}

\section{Task Formalization}
Continual relation extraction aims to continually learn new relations from a series of $n$ tasks
$\mathbb{T}=\left(\mathcal{T}^1, \mathcal{T}^2, \ldots, \mathcal{T}^n\right)$. where the $k$-th task $\mathcal{T}^k$ has its own relation set $R^k$, and each relation $r \in R^k$ owns several instances $\bm{x}$, $\bm{x} = [x_1, \ldots, e_h, \ldots, e_t, \dots, x_{|\bm{x}|]}]$ which represents a natural language sentence with head entity $e_h$ and tail entity $e_t$. The samples from $\mathcal{T}^k$ are divided into 
training set $D_{train}^k$, validation set $D_{valid}^k$, and test set $D_{test}^k$. The relations from different tasks are disjoint $\cap_{i=1}^n R^i = \emptyset$.
Note that, once finishing task $\mathcal{T}^k$, 
the samples of $\mathcal{T}^k$ are no longer available for future training.
The model after training $\mathcal{T}^k$ 
will be evaluated on the test set of all seen relations.

To align with the real situation where labeled samples of each relation are often scarce, we focus on CRE in the few-shot setting following the work of \citet{qin2022continual}.
For the first task $\mathcal{T}^1$, samples are sufficient, and the subsequent tasks $\mathcal{T}^2, \mathcal{T}^3, \ldots, \mathcal{T}^n$ are all the few-shot.
Following \cite{chen2023consistent}, we set $N$ as the relation number of each few-shot task and $K$ as the sample number of each relation,
so the continual few-shot task can be called \textit{continual} $N$-\textit{way} $K$-\textit{shot} task.

Following previous works \cite{han2020continual, qin2022continual, wang2023serial}, 
we set a memory $\mathcal{M}=\left\{\mathcal{M}^1, \mathcal{M}^2, \ldots, \mathcal{M}^n\right\}$ with size $L$,
which stores $L$ typical samples for each relation from previous tasks.
When training on $\mathcal{T}^k$, the memory $\hat{\mathcal{M}}^{k-1} = \cup_{i=1}^{k-1} \mathcal{M}^i$ can be accessed by the model. In the few-shot setting, we just store one sample per relation ($L=1$) in the memory.

\section{Method}
In this section, we detail the components of the proposed method and the training procedure.
\subsection{Framework Overview}
As shown in figure \ref{framework}, our framework can be divided into three modules:
(1) \textbf{prompt representation}, 
(2) \textbf{contrastive learning}, and 
(3) \textbf{memory augmentation}.
Prompt representation module introduces the prompt template and the encoding method, which makes downstream tasks more similar to pre-training tasks of PLMs, so that the model is better adapted to new tasks, detailed in Section~\ref{4.2}.
Contrastive learning module proposes a margin-based contrastive learning objective and introduces how to train the model during current task and memory replay. After two contrastive training, model tends to learn a more uniform feature distribution, thus reducing overfitting, detailed in Section~\ref{4.3}.
Memory augmentation module selects typical samples from current tasks and then generates extra samples by ChatGPT for data augmentation, it further mitigates overfitting caused by sparse data, see Section~\ref{4.4}.

\subsection{Prompt Representation}\label{4.2}

Vanilla prompt engineering consists of two parts,
namely template and verbalizer. The template reformulates the original input as a cloze-style phrase by adding a set of prompt tokens and one \texttt{[MASK]} token, and the verbalizer maps each task label to the corresponding textual token. By predicting the \texttt{[MASK]} token as label verbalization,
the model can determine the corresponding label of the input example. Different from vanilla prompt engineering, our prompt framework only includes template without verbalizer. By adding a prompt template to input sentences, the downstream tasks are more similar to the pre-training task of PLMs, therefore PLMs can understand well what to do and perform better on downstream tasks.

\paragraph{Prompt template}
We design a semi-automated continuous template $T$,
which combines entity information and learnable tokens as prompts
rather than hard prompt with only handcrafted tokens 
or soft prompt with only learnable tokens \cite{lester2021power}.
Hard prompt likes 
``$\bm{x}$. \textit{The relation between $e_h$ and $e_t$ is} \texttt{[MASK]}.'', 
which needs different sophisticated expert knowledge for specific tasks,
and soft prompt might be hard to converge with sparse data.
Specifically, 
for an input sentence $\bm{x}$ with two entities $e_h$ and $e_t$,
the template is as follows:
\begin{align}
  T(\bm{x}) = \; &\bm{x}. \left[v_{0:n_0-1}\right] \bm{e_h} \left[v_{n_0:n_1-1}\right] \rm{\left[\texttt{MASK}\right]} \notag\\
  &\left[v_{n_1:n_2-1}\right] \bm{e_t} \left[v_{n_2:n_3-1}\right].  
\end{align}
where $\left[ v_i \right]$
refers to the $i$-th learnable continuous token,
and $n_i$ is the length of token phrases.
We add a special token $\rm{\left[\texttt{MASK}\right]}$ in the template as the representation of the relation between $\bm{e_h}$ and $\bm{e_t}$.

\paragraph{Encoding}
Given a encoding model $E$ and a templated sentence $T \left( \bm{x} \right)$ as input,
we map the $T \left( \bm{x} \right)$ into  a sequence of continuous vectors:
\begin{align}
  E_{mb} \left(T\left(\bm{x}\right)\right) & = \; e\left( \bm{x} \right), h_0, \ldots, h_{n_0-1}, e\left( \bm{e_h} \right), \notag \\
  &h_{n_0}, \ldots, h_{n_1-1}, e\left(\left[ \texttt{MASK} \right]\right), h_{n_1}, \notag \\
  &\ldots, h_{n_2-1}, e\left( \bm{e_t} \right), h_{n_2}, \ldots, h_{n_3-1} 
\end{align} 
where $E_{mb}\left( \cdot \right)$ is the embedding function, $e\left( \right)$ is the embedding layer of encoding model $E$,
$h_i \in \mathbb{R}^{d_1}$ are learnable vectors, 
$d_1$ is the embedding dimension of $E$,
and $0\,\leq\,i\,<\,n_{3}$.
Then the embeddings are fed into the encoding model $E$
and get the hidden representations $\bm{m}$ of the sentence:
\begin{align}
  \bm{m} = E_{nc} \left( E_{mb} \left( T\left( \bm{x} \right) \right) \right) 
\end{align}
where $E_{nc}\left( \cdot \right)$ represents the hidden layers of $E$.
$\bm{m} \in \mathbb{R}^{d_2}$ refers the hidden representation of $\rm{\left[\texttt{MASK}\right]}$,
and $d_2$ is the hidden dimension of $E$.

\subsection{Contrastive Learning}\label{4.3}

We design a novel margin-based contrastive learning to gain discriminative representations and focus more on hard samples, thus alleviating overfitting. It is worth noting that by doing this, the model does not need the verbalizer module in vanilla prompt engineering, which saves the labor required for prompt engineering and makes the method more general.
Instead of linear classifier with extra parameters in previous works \cite{cui2021refining,qin2022continual}, our method predicts relations based on parameter-free metric method, which is more suitable for incremental-class problems.

\paragraph{Margin-based contrastive learning objective}
Typically, let $\bm{z_i}$ be PLMs' output hidden vector after normalization and $s_{i,p} = \bm{z_i} \cdot \bm{z_p}$ denotes positive pair, $s_{i,n} = \bm{z_i} \cdot \bm{z_n}$ denotes negative pair, the MCL loss can be defined as follows:

\begin{align}
  \mathcal{L}_{\mathrm{MCL}}\left(i\right)= \sum_{p \in P(i)} 
  \log \frac{\exp \left(\alpha_{i,p} \cdot s_{i,p} / \tau\right)}{\mathcal{Z}\left(i\right)} , 
\end{align}
where $P(i)$ denotes the positive set with $i$, 
$\tau$ is the temperature constant
and $\mathcal{Z}\left(i\right)$ is as follows: 
\begin{align} \label{eqz}
  \mathcal{Z}(i)\!=\! \sum_{p \in P(i)} \! \exp \left(\alpha_{i, p} \frac{s_{i, p}}{\tau} \right) 
  \!+\! \sum_{n \in N(i)} \! \exp \left(\alpha_{i, n} \frac{s_{i, n}}{\tau} \right) 
\end{align}
where $N(i)$ is the negative set with $i$, 
$\alpha_{i,p}$ and $\alpha_{i,n}$ are relaxation factors that control the relaxation of the decision boundaries.
\begin{align}
  \alpha_{i,p} = m + k \cdot s_{i,p}, \;
  \alpha_{i,n} = 1 - m + k \cdot s_{i,n} 
\end{align}
where $k$ is normalization constant depends on $s_{i,p}$ and $s_{i,n}$, $m$ is the margin factor 
which expects $k \cdot s_{i,p} > 1-m$ and $k \cdot s_{i,n} < m$.

As we can see, MCL objective can make models pay more attention to hard samples and less attention to easy ones, thus mitigating overfitting problem.
It can also make learned distribution more uniform and further alleviate catastrophic forgetting.

\paragraph{Current task training}
Given a new task $\mathcal{T}^k$, 
we perform training process with MCL loss on each batch $B$.
According to \cite{khosla2020supervised}, 
supervised contrastive loss performs better with big batch size,
but it's costly to have a large memory,  
so we set an additional memory $S$ for contrastive training.
Specifically, 
the sentences of current training set $D_{train}^k$ are encoding as features $\bm{z}$ 
and stored in a bucket $\mathcal{C}^k$.
Then, 
for i-th instance $\bm{x}$ in $B$, 
we randomly select partial features from $\mathcal{C}^k$ 
to form a temporary contrastive features set $S_i$.
Finally, 
the contrastive loss of i-th instance $\bm{x}$ can be calculated in $S_i$.
The batch MCL loss is as follows:
\begin{align} \label{eq7}
  \mathcal{L}_{\mathrm{MCL}}=\sum_{i \in I} \frac{-1}{|P(i)|} \sum_{p \in P(i)} 
  \log \frac{\exp \left(\alpha_{i,p} \cdot s_{i,p} / \tau\right)}{\mathcal{Z}\left(i\right)} , 
\end{align}
where $I = \left\{1,2, \ldots , |B|\right\}$, and positive set $P(i) = \left\{p \in S_i : y_p = y_i \right\}$, negative set $N(i) = \left\{n \in S_i : y_n \neq y_i \right\}$.
After backpropagating the gradient of loss on each batch, we update the corresponding features in the bucket $\mathcal{C}^k$:
\begin{align} \label{eq8}
  \mathcal{C}^k [ \hat{I} ] \gets {\left\{\bm{z}_i\right\}}_{i=1}^{|B|}, 
\end{align}
where $\hat{I}$ is the corresponding index set of this batch of samples in $\mathcal{C}^k$.
\paragraph{Memory replay}
The new task training process can make the model concentrate more on new knowledge while forgetting old learned knowledge, especially in low-resource scenarios.
To alleviate forgetting, the second training is performed in the memory to consolidate old knowledge.
Specifically, we generate additional samples to augment the memory set (refer to section \ref{secdataaug}) and then perform memory training with MCL loss again, which can not only allow the model to review learned knowledge but also prevent the model from overfitting.

\subsection{Memory Augmentation}\label{4.4}
After current task training, We first select typical samples of current training data to store in memory, then design the elaborate prompt to guide ChatGPT to generate diverse samples for memory augmentation.

\paragraph{Representative memory sampling}
Inspired by previous works \cite{han2020continual}, after training for current task $\mathcal{T}^k$, we apply K-means algorithm \cite{likas2003global} to select typical samples for each relation and store them in the memory $\hat{\mathcal{M}}$.
Specifically, we first get features $\bm{z}_i$ of i-th $\bm{x}$ on $D_{train}^k$, and for each relation $r \in R^k$, we use K-means algorithm to cluster features ${\left\{z_i\right\}}_{i=0}^{|R^k|}$ into $L$ clusters.
Then, for each cluster, the nearest sample closest to the centroid is chosen as the typical sample which is stored in the memory $\hat{\mathcal{M}}$.

\paragraph{Prompt data augmentation} \label{secdataaug}
To utilize the rich knowledge of LLMs, we design elaborate prompt to stimulate LLMs' powerful language generation ability to generate relevant examples.
We choose GPT-3.5 as our generation model. For every historical relation $r$ in $\hat{R}^k$, we select one typical sample from memory $\hat{\mathcal{M}}^k$ 
as the instance and construct a prompt input including task instruction, semantic relation explanation (in purple) and demonstrations (in blue) like In-context learning \cite{dong2022survey}. Here is an example of prompting ChatGPT to generate samples with the relation \textit{``founded by''}:
\begin{center}
\fcolorbox{black}{gray!10}{\parbox{.96\linewidth}{
\textbf{Prompt:}

\textcolor{black}{One sample in relation extraction datasets consists of a relation, a context, a head and tail entity. The head entity has the relation with the tail entity.}
\textcolor{purple}{\textit{Relation \textbf{founded by} means an organization was found by a person.}}

\textcolor{black}{Here is an example:}

\textcolor{blue}{Relation: founded by}

\textcolor{blue}{Context: Steve Jobs is the co-founder of Apple Inc.}

\textcolor{blue}{Head Entity: Steve Jobs}

\textcolor{blue}{Tail Entity: Apple Inc.}

Please generate $n$ samples for relation \textit{\textbf{founded by}}:
}}
\end{center}
Then it is used to ask ChatGPT to generate $g$ diverse samples with relation $r$.
For the output, we parse the text to get structured data as the generated training data $\mathcal{A}$.
Finally, memory samples $\hat{\mathcal{M}}^k$ 
are combined with generated data $\mathcal{A}$ 
to construct a new training set for subsequent memory replay.
\subsection{Training Procedure}\label{4.5}
First, the parameters of prompt template $\theta^1$ and encoding model $\theta^2$ are initialized with the parameters trained on the last task.
Then, there are two main training steps:
(1) \textbf{Current task training}: 
current task training set $D_{train}^k$ is encoded with prompt template $\theta^1$,
and then fed into the encoding model $\theta^2$ for training 
by MCL loss $\mathcal{L}_{\mathrm{MCL}}$
(2) \textbf{Memory replay}:
after training current task,
we select typical samples for each relation  
and store them in the memory $\hat{\mathcal{M}}$.
Then, ChatGPT is applied to generate relevant samples 
with the same relation in memory $\hat{\mathcal{M}}$ 
to augment training set for subsequent memory replay.
Finally, historical memory samples are trained together with generated samples via the MCL loss $\mathcal{L}_{\mathrm{MCL}}$, allowing the model to recall forgotten knowledge without overfitting due to sparse data.

\subsection{Relation Prediction} \label{secrelationpre}
To leverage the discriminative feature distribution after contrastive training,
Nearest-Class-Mean (NCM) classifier is adopted to predict the relations in the test phase.
Given a historical memory $\hat{\mathcal{M}}^n$,
we perform forward propagation with trained encoding model $E$
to get the features of all relations 
and average features for each relation to calculate prototypes of relations.
Then we get the feature of test sample $\bm{x}$
to compute L2 pairwise distance with all prototypes of seen relations
and get the nearest prototype $p_r$ with relation label:
\begin{align}
  &p_r = \frac{1}{L} \sum_{i=0}^{L} \bm{E} \left(\bm{\hat{x}}_i^r\right), \notag \\
  &y^*=\underset{r=1, \ldots, n}{\operatorname{argmin}}{\left\| \bm{E}\left(\bm{x}\right)-p_r\right\|}_2 
\end{align}
where $L$ is the memory size per relation,
$\bm{\hat{x}}_i^r$ denotes the samples with label $y_r$ in memory $\hat{\mathcal{M}}^n$.
$y^*$ is the predicted relation label.

\begin{table*}
  \centering
  \small
  \scalebox{0.9}{
  \begin{tabular}{lcccccccc}
    \toprule
    \multicolumn{9}{c}{\textbf{FewRel} (\textit{10-way 5-shot})}\\
    \midrule
    \textbf{Method} & $\bm{\mathcal{T}^1}$ & $\bm{\mathcal{T}^2}$ & $\bm{\mathcal{T}^3}$ & $\bm{\mathcal{T}^4}$ & $\bm{\mathcal{T}^5}$ & $\bm{\mathcal{T}^6}$ & $\bm{\mathcal{T}^7}$ & $\bm{\mathcal{T}^8}$ \\ 
    \midrule
    Finetune & 94.58 & 41.67 & 26.42 & 20.79 & 16.63 & 13.09 & 11.27 & 9.78 \\
    Joint-train & \textbf{95.07} & \textbf{88.16} & \textbf{83.08} & \textbf{79.67} & \textbf{77.85} & \textbf{75.11} & \textbf{72.62} & \textbf{70.33} \\
    \midrule
    EMAR+ACA\textsuperscript{\ddag} \cite{wang2022learning} & 94.75 & 78.33 & 69.01 & 67.17 & 65.55 & 61.77 & 60.04 & 58.48 \\
    InfoCL\textsuperscript{\ddag} \cite{song2023infocl} & 95.38 & 78.92 & 72.63 & 69.05 & 66.75 & 63.36 & 60.65 & 58.90 \\
    RP-CRE\textsuperscript{\dag} \cite{cui2021refining} & 93.97 & 76.05 & 71.36 & 69.32 & 64.95 & 61.99 & 60.59 & 59.57 \\
    CRL\textsuperscript{\dag} \cite{zhao2022consistent} & 94.68 & 80.73 & 73.82 & 70.26 & 66.62 & 63.28 & 60.96 & 59.27 \\
    CRECL\textsuperscript{\dag} \cite{hu2022improving} & 93.93 & 82.55 & 74.13 & 69.33 & 66.51 & 64.60 & 62.97 & 59.99 \\
    ERDA\textsuperscript{\dag} \cite{qin2022continual} & 92.43 & 64.52 & 50.31 & 44.92 & 39.75 & 36.36 & 34.34 & 31.96 \\
    SCKD\textsuperscript{\ddag} \cite{wang2023serial} & 94.75 & 82.56 & 75.98 & 72.04 & 70.53 & 67.02 & 64.73 & 62.87 \\
    \textbf{CPL (ours)} & \underline{94.87} & \underline{85.14} & \underline{78.80} & \underline{75.10} & \underline{72.57} & \underline{69.57} & \underline{66.85} & \underline{64.50} \\
    \midrule
    \midrule
    \multicolumn{9}{c}{\textbf{TACRED} (\textit{5-way 5-shot})}\\
    \midrule
    \textbf{Method} & $\bm{\mathcal{T}^1}$ & $\bm{\mathcal{T}^2}$ & $\bm{\mathcal{T}^3}$ & $\bm{\mathcal{T}^4}$ & $\bm{\mathcal{T}^5}$ & $\bm{\mathcal{T}^6}$ & $\bm{\mathcal{T}^7}$ & $\bm{\mathcal{T}^8}$ \\ 
    \midrule
    Finetune & 88.12 & 23.26 & 17.23 & 16.30 & 14.19 & 9.64 & 9.47 & 7.90 \\
    Joint-train & \textbf{88.40} & \textbf{83.65} & \textbf{74.40} & \textbf{71.36} & \textbf{65.61} & \textbf{64.01} & \textbf{61.13} & \textbf{58.26} \\
    \midrule
    EMAR+ACA\textsuperscript{\ddag} \cite{wang2022learning} & 88.06 & 71.55 & 63.57 & 51.84 & 46.03 & 46.31 & 42.55 & 39.78 \\
    InfoCL\textsuperscript{\ddag} \cite{song2023infocl} & 87.45 & 75.35 & 65.94 & 59.10 & 52.24 & 49.42 & 44.52 & 40.37 \\    
    RP-CRE\textsuperscript{\dag} \cite{cui2021refining} & 87.32 & 74.90 & 67.88 & 60.02 & 53.26 & 50.72 & 46.21 & 44.48 \\
    CRL\textsuperscript{\dag} \cite{zhao2022consistent} & \underline{88.32} & 76.30 & 69.76 & 61.93 & 54.68 & 50.92 & 47.00 & 44.27 \\
    CRECL\textsuperscript{\dag} \cite{hu2022improving} & 87.09 & 78.09 & 61.93 & 55.60 & 53.42 & 51.91 & 47.55 & 45.53 \\
    ERDA\textsuperscript{\dag} \cite{qin2022continual} & 81.88 & 53.68 & 40.36 & 36.17 & 30.14 & 22.61 & 22.29 & 19.42 \\
    SCKD\textsuperscript{\ddag} \cite{wang2023serial} & 88.16 & 78.89 & 71.68 & 64.99 & 61.13 & 57.92 & 53.18 & 51.11 \\
    \textbf{CPL (ours)} & 86.27 & \underline{81.55} & \underline{73.52} & \underline{68.96} & \underline{63.96} & \underline{62.66} & \underline{59.96} & \underline{57.39} \\ 
    \bottomrule
  \end{tabular}
  }
  \caption{Main results on FewRel and TACRED in 5-shot setting. \dag \, are reported in \cite{wang2023serial}, \ddag \, are we re-running the origin code.
  The \textbf{best results} are in bold, and the \underline{second-highest} are underlined.}
  \label{main}
\end{table*}

\section{Experiments}
In this section, we conduct experiments on two RE datasets and give a further analysis of the results.

\subsection{Experiments Setup}
\paragraph{Datasets}
We conduct experiments on two RE datasets: FewRel \citeplanguageresource{han2018fewrel} and TACRED \citeplanguageresource{zhang2017position}.
For \textbf{FewRel}, we follow \citet{wang2023serial}, and adopt the version of 80 publicly released relations, and split them equally into 8 tasks.
For the first task $\mathcal{T}^1$, we select 100 instances per relation, while the subsequent tasks $\left\{ \mathcal{T}^i \right\}_{i=2}^8$ are few-shot tasks with only 5 and 10 instances per relation for training which called \textit{10-way 5-shot} and \textit{10-way 10-shot}, respectively.
For \textbf{TACRED}, we filter out \textit{``no\_relation''} and divide the remaining 41 relations into 8 tasks.
The first task $\mathcal{T}^1$ has 6 relations and 100 instances per relation, and subsequent tasks $\left\{ \mathcal{T}^i \right\}_{i=2}^8$ have 5 relations with 5 or 10 instances per relation. 

\paragraph{Evaluation metrics}
We evaluate the model's performance by the average accuracy of all tasks.
After training on task $\mathcal{T}^k$, the model is evaluated on $\hat{D}_{test}^k = \cup_{i=0}^k D_{test}^i$ to get the overall accuracy of all seen relations.
We measure the average performance across six rounds of experiments.

\paragraph{Baselines}
We compare with some strong CRE models published in recent years:
\textbf{RP-CRE} \cite{cui2021refining} introduces a novel attention-based memory module to refine subsequent sample embeddings. \textbf{CRL} \cite{zhao2022consistent} and \textbf{CRECL} \cite{hu2022improving} adopt contrastive learning to ensure the feature space is more distinguishable. \textbf{EMAR+ACA} \cite{wang2022learning} propose an adversarial class augmentation mechanism using \textbf{EMAR} \cite{han2020continual} as the backbone. \textbf{InfoCL} \cite{song2023infocl} focus on distinguishing analogous classes. \textbf{ERDA} \cite{qin2022continual} and \textbf{SCKD} \cite{wang2023serial} mainly focus on CRE in few-shot, they use additional memory to alleviate forgetting and data augmentation to mitigate overfitting. 

We set two borderlines:
\textbf{Finetune} trains models sequentially without memory set. It faces serious catastrophic forgetting and serves as the lower bound.
\textbf{Joint-training} stores all samples of previous tasks in memory, and trains models with all data for every new task, which can be regarded as the upper bound. 

\paragraph{Implementation details}
We implement our framework based on PyTorch 1.7.0 \cite{paszke2019pytorch} and Huggingface's Transformers 4.10.0 \cite{wolf2020transformers}. 
For a fair comparison, BERT-base-uncased \cite{devlin2019bert} is adopted as our encoding model.
We choose GPT-3.5-turbo \cite{openai2023chat35} for sample generation.
We set the random seeds identical to \citet{wang2023serial,qin2022continual} so that the task order is exactly the same. Following \citet{wang2023serial}, we employ the \textit{``strict''} evaluation rule to put all seen relation labels as candidates when predicting, rather than \textit{``loose''} rule which only picks 10 labels as candidates \cite{qin2022continual}.
For memory size, we set $L=1$, i.e. there is only one sample per relation.
See Appendix \ref{moreparams} for more details.

\subsection{Results and Analysis}
\subsubsection{Main Results}
Table \ref{main} presents the overall results of the 5-shot setting on FewRel TACRED (See Appendix \ref{10shotresult} for 10-shot results).
We have the following findings:

(1) Our CPL outperforms all CRE models and achieves SOTA performance on both RE datasets.
Specifically, we outperform the second-best model SCKD by 1.63\% and 6.28\% on FewRel and TACRED, respectively.
Notably, we achieve significantly the best performance on the final task of TACRED, even though the performance on the first task is not very good, which proves that the proposed CPL better handles the catastrophic forgetting and overfitting problems.
(2) Observing the performance of two borderlines, we find finetune model leads to rapid drops in average accuracy, which shows catastrophic forgetting and overfitting in CFRE tasks.
Particularly, we achieve close performance to the joint-training model on TACRED with only one memory sample per relation.
However, there is still a five-point gap behind the joint train on FewRel due to the bias caused by very few memory samples. 
(3)Compared to data augmentation methods ERDA and SCKD, CPL achieves significant improvements, which demonstrates that our method can generate more diverse samples that are closer to the true distribution.
Note that ERDA gains the worst performance in our setting, the reason is that the extra data they import may contain noise and bias, and loose evaluation metrics they used are also ill-considered.
(4) Compared to contrastive learning methods CRL and CRECL, CPL outperforms them by a large margin on both datasets, which demonstrates that our proposed margin-based contrastive learning can better alleviate overfitting and is more suitable in low-resource scenarios.

\subsubsection{Ablation Study}

\begin{table}[t]
  \centering
  \tabcolsep=0.1cm
  \resizebox{\linewidth}{!}{
  \begin{tabular}{lcccccccc}
    \toprule
    \textbf{FewRel} & $\bm{\mathcal{T}^1}$ & $\bm{\mathcal{T}^2}$ & $\bm{\mathcal{T}^3}$ & $\bm{\mathcal{T}^4}$ & $\bm{\mathcal{T}^5}$ & $\bm{\mathcal{T}^6}$ & $\bm{\mathcal{T}^7}$ & $\bm{\mathcal{T}^8}$ \\ 
    \midrule
    CPL & 94.87 & 85.14 & 78.80 & 75.10 & 72.57 & 69.57 & 66.85 & 64.50 \\
    \enspace w.o. pro. & 93.78 & 78.86 & 69.65 & 63.97 & 60.89 & 56.73 & 53.83 & 51.09 \\
    \enspace w.o. MCL  & 94.97 & 84.05 & 77.64 & 73.32 & 70.21 & 66.76 & 63.71 & 61.78 \\
    \enspace w.o. gen  & 94.73 & 82.62 & 76.16 & 73.38 & 70.51 & 66.98 & 65.30 & 63.78\\
    \enspace w.o. all  & 93.20 & 76.46 & 66.14 & 61.49 & 57.97 & 53.72 & 50.09 & 48.29 \\
    \midrule
    \textbf{TACRED} & $\bm{\mathcal{T}^1}$ & $\bm{\mathcal{T}^2}$ & $\bm{\mathcal{T}^3}$ & $\bm{\mathcal{T}^4}$ & $\bm{\mathcal{T}^5}$ & $\bm{\mathcal{T}^6}$ & $\bm{\mathcal{T}^7}$ & $\bm{\mathcal{T}^8}$ \\ 
    \midrule
    CPL & 86.27 & 81.55 & 73.52 & 68.96 & 63.96 & 62.66 & 59.96 & 57.39 \\
    \enspace w.o. pro. & 86.30 & 75.14 & 64.27 & 59.44 & 54.11 & 48.87 & 44.74 & 42.61 \\
    \enspace w.o. MCL  & 86.02 & 81.37 & 75.46 & 71.46 & 64.60 & 63.26 & 57.53 & 54.75 \\
    \enspace w.o. gen  & 86.14 & 79.61 & 70.01 & 65.24 & 60.16 & 57.98 & 52.64 & 50.63\\
    \enspace w.o. all  & 86.77 & 76.10 & 61.05 & 55.01 & 50.83 & 44.71 & 39.76 & 38.06 \\
    \bottomrule
  \end{tabular}
  }
  \caption{Ablation study on components of CPL.}
  \label{ablation}
\end{table}

We conduct ablation experiments on the FewRel 10-way 5-shot and TACRED 5-way 5-shot to verify the effectiveness of each component in our method.
(a) w.o. pro. denotes without prompt template. Instead, we use entity markers following previous works \cite{cui2021refining, zhao2022consistent, qin2022continual, hu2022improving, wang2023serial}.
(b) w.o. MCL means using supervised contrastive loss \cite{zhao2022consistent} 
instead of our margin-based contrastive loss.
(c) w.o. gen represents without generated samples for memory replay.
(d) w.o. all means without all of the above components.

As shown in table \ref{ablation}, prompt representation brings a boost of 13.41\% and 14.78\% after training all tasks, demonstrating the powerful effectiveness of prompt learning for alleviating catastrophic forgetting.
MCL loss can also contribute to 2.27\% and 2.26\% improvement in the 5-shot setting, which shows our margin-based contrastive learning helps models mitigate overfitting caused by sparse data. 
Additionally, we observe a steady improvement with data augmentation, which represents ChatGPT effectively generates various samples to help the model learn more realistic distributions.

\subsubsection{Analysis of Prompt Representation}
\begin{figure}[t]
    \centering
    \includegraphics[width=0.45\textwidth]{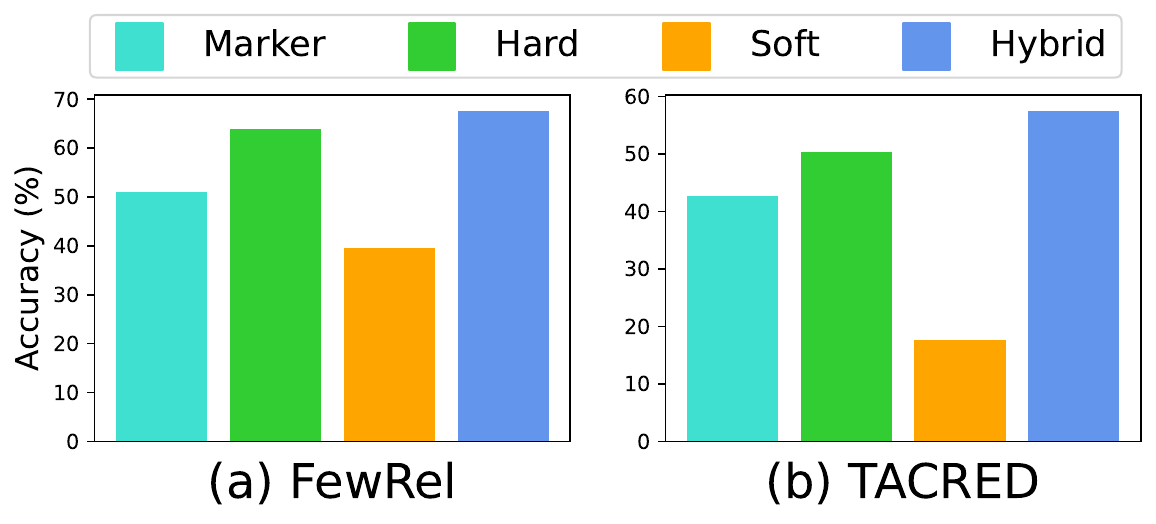}
    \caption{Results of different prompt formats.}
    \label{prompt}
\end{figure}

To verify the effectiveness of proposed prompt representation method, we conduct comparison experiments with four encoding templates in table \ref{template}: entity marker, hard prompt, soft prompt, and hybrid prompt template.
Figure \ref{prompt} shows the average accuracy after final task training which represents the degree of forgetting.
It can be observed that hybrid prompt outperforms entity marker extensively used in previous works by a big gap, demonstrating the effectiveness of proposed prompt template.
The reason is that we combine entity knowledge for RE and make downstream tasks approximate the tasks during pre-training phase of PLMs, hence, the current task training can activate underlying knowledge of PLMs and further alleviate forgetting.
Notably, hard prompt is elaborately handcrafted for RE tasks, which contains domain expert knowledge, but even so, our hybrid prompt still outperforms it distinctly, 
which shows that our hybrid prompt can learn the most suitable template for current tasks with only several samples. The reason for the poor performance of soft prompt is that randomly initialized vectors are difficult to converge with only a few samples, as reported by \citet{gu2022ppt}.

\begin{table}[t]
  \centering
  \begin{tabularx}{0.45\textwidth}{lX}
    \toprule
    \multicolumn{2}{l}{Template} \\
    \midrule
    1 & $x_0$ \ldots [$E_0$] $\bm{e_h}$ [$E_1$] \ldots [$E_2$] $\bm{e_t}$ [$E_3$] \ldots $x_n$. \\
    2 & $\bm{x}$. \textit{The relation between $\bm{e_h}$ and $\bm{e_t}$ is} [MASK]. \\
    3 & $\bm{x}$. [$v_0$] [$v_1$] [MASK] [$v_2$] [$v_3$]. \\
    4 & $\bm{x}$. [$v_0$] $\bm{e_h}$ [$v_1$] [MASK] [$v_2$] $\bm{e_t}$ [$v_3$]. \\
    \bottomrule
  \end{tabularx}
  \caption{Templates of different prompt encoding.}
  \label{template}
\end{table}

\subsubsection{Visualization of Contrastive Learning}
\begin{figure}[t]
  \centering
  \includegraphics[width=0.46\textwidth]{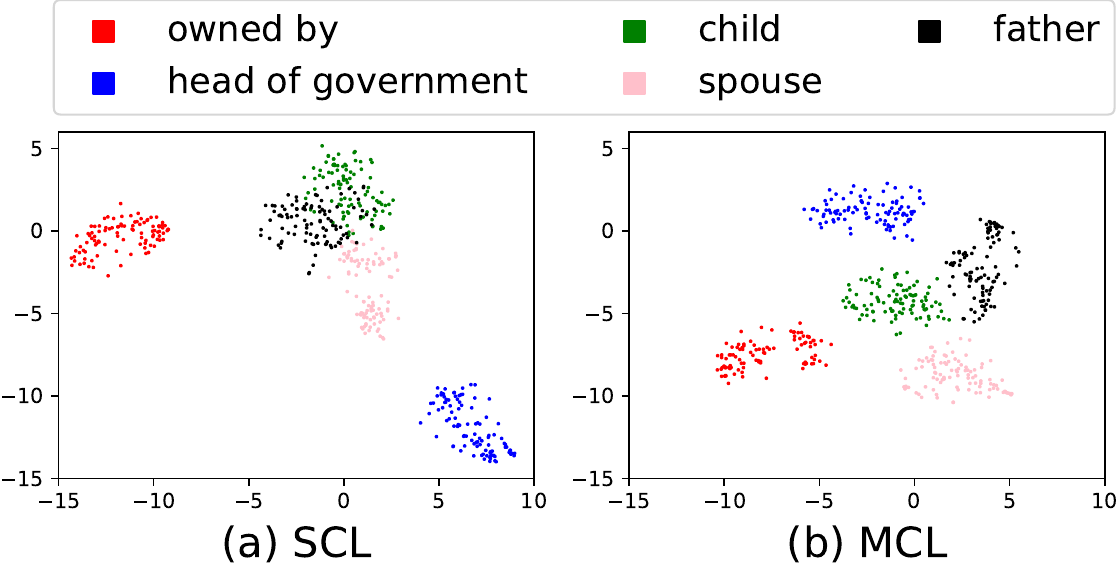}
  \caption{t-SNE plots of instance embeddings trained with SCL and MCL. }
  \label{tsne}
\end{figure}

In order to study the effects of proposed margin-based contrastive objective in low resource scenarios,
we use t-SNE \cite{van2008visualizing} to visualize the feature space of test data after current task training. 
Figure \ref{tsne} illustrates the results of supervised contrastive loss (SCL) and our MCL loss. As we can see, the feature space of MCL is more uniform than SCL, and samples with different relations are discriminative.
As for SCL, though samples with easy-distinguish relations like \textit{``owned by''} and \textit{``head of government''} have obvious boundaries, some samples with similar relations as \textit{``child''} and \textit{``father''} are difficult to distinguish.
This is because SCL focuses more on easy pairs and less on hard ones, making it difficult to distinguish similar relations, especially in the few-shot setting.

\begin{figure}[t]
  \centering
  \includegraphics[width=0.45\textwidth]{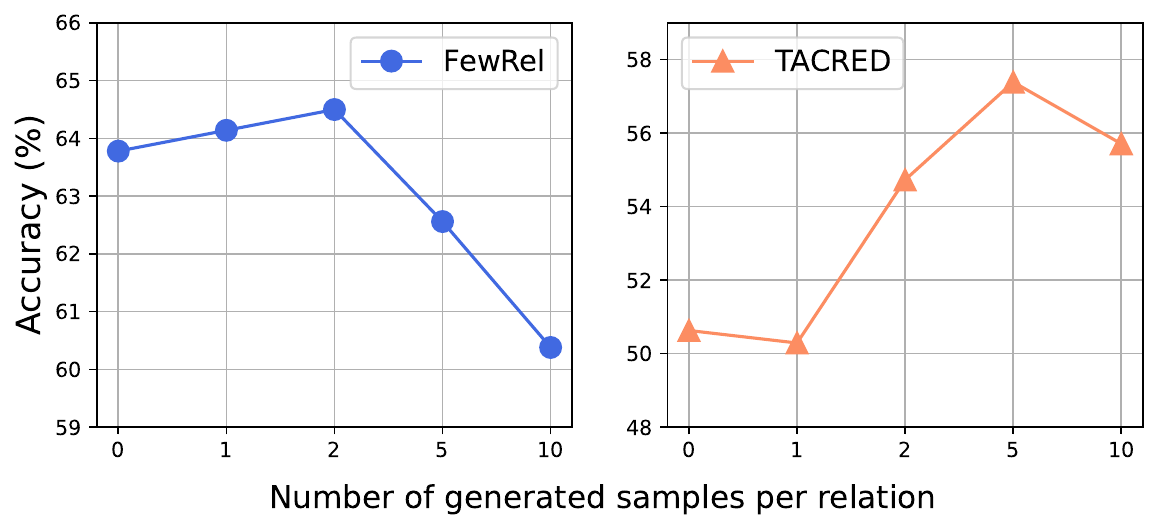}
  \caption{Results with different generated samples.}
  \label{generate}
\end{figure} 

\subsubsection{
Analysis of Memory Augmentation through LLMs
} \label{secdatagen}
To analyze the effect of memory augmentation through LLMs on results, we conduct experiments with different numbers of generated samples in the 5-shot setting, where one sample per relation is stored in memory ($L=1$). 
As shown in Figure \ref{generate}, data generation with LLMs can yield better performance in the few-shot setting.
However using more generated data can only boost RE performance to a certain extent, not continuously better, and even cause a performance degradation with on FewRel.
This is because the samples generated by LLMs may be noisy and not diverse enough.
We further conduct a detailed case study to show the real inputs and corresponding outputs of ChatGPT, as presented in table \ref{case}.
We combine relation name and description (in purple italics) as prompt input, output 1 \textit{with example} in input while output 2 \textit{without example}.
We observe that LLMs can generate the correct sample with the relation \textit{``child''} when with the example, but generate the sample with the wrong relation \textit{``siblings''} between entities when without example.
It shows some relations can not be understood well by LLMs only with relation names and descriptions, examples are sometimes necessary for LLMs. 
More results refer to Appendix \ref{gencase}.

\begin{table}
  \centering
  \begin{tabularx}{0.48\textwidth}{X}
    \toprule
      \textbf{Prompt:} 
      One sample in relation extraction datasets consists of a relation, 
      a context, a head and tail entity. The head entity has the relation with the tail entity.
      \textit{\color{purple}{Relation \textbf{child} means subject has object as child}}.\\
      Here is an example: \\
      Relation: child \\
      Context: \textbf{\color{red}{Pierre-Suzanne-Augustin Cochin}} was the father of 
      the Catholic politician \textbf{\color{blue}{Denys Cochin}} 
      and the grandfather of the historian Augustin Cochin. \\
      Head Entity: Pierre-Suzanne-Augustin Cochin \\
      Tail Entity: Denys Cochin \\
      Please generate $n$ samples for relation \textbf{child}:
      \\
    \midrule
      \textbf{Output 1: (with example in input)} \\
      Relation: child \\
      Context: \textbf{\color{red}{John Smith}} is the father of \textbf{\color{blue}{Emily Smith}} and the husband of Sarah Smith. 
      Emily is a talented musician, while Sarah is a painter. \\
      Head Entity: John Smith \\
      Tail Entity: Emily Smith \\
    \midrule
      \textbf{Output 2: (without example in input)} \\
      Relation: child \\
      Context: \textbf{\color{red}{John}} and \textbf{\color{blue}{Mary}} are \textbf{siblings}. 
      John is the eldest child in the family. \\
      Head Entity: John \\
      Tail Entity: Mary \\
    \bottomrule
  \end{tabularx}
  \caption{Case of the inputs and outputs of memory augmentation.}
  \label{case}
\end{table}

\section{Conclusion}
In this paper,  we propose CPL, which leverages prompt learning and contrastive learning to empower pre-trained language models to better continual few-shot relation extractors.  To alleviate catastrophic forgetting, we explore using prompt learning to activate generalized knowledge of PLMs that can easily adapt to old and new relations. To mitigate overfitting, we design a novel margin-based contrastive learning objective and apply ChatGPT to generate diverse samples for memory augmentation. Extensive experiments indicate that CPL enables PLMs to better mitigate catastrophic forgetting and overfitting in low-resource scenarios. In the future, we are going to study how to design better prompts to guide LLMs to generate more reliable samples.

\section{Limitations}
We argue that the main limitations of our work are mainly twofold:
(1) Time efficiency. As the model continues to learn new tasks, more different relation samples are held in memory, and more augmented data are generated by the memory samples.
Therefore, with the continuous accumulation, the sample of subsequent tasks will be more and more, and the training time will be more and more. In the future, we will explore more time-efficient data augmentation strategies. (2) Stability of results. Due to the unstable content generated by ChatGPT, augmented samples are inconsistent in some experiments with the same input, even when we had set the temperature in OpenAI to 0 for consistency. Therefore, under the same setting, experimental results are not completely consistent each time. We tried our best to ensure the validity of the experimental results by taking the average value of six rounds of experiments.

\section{Acknowledgments}
We would like to thank the anonymous reviewers, our meta-reviewer, and senior area chairs for
their thoughtful comments and support on this work.
This work was supported in part by the National Key Research and Development Program of China under Grant 2022YFF0902701, in part by the National Natural Science Foundation of China under Grants U21A20468, 62372058, U22A201339, in part by the Fundamental Research Funds for the Central Universities under Grant 2020XD-A07-1.

\nocite{*}
\section{Bibliographical References}
\bibliographystyle{lrec-coling2024-natbib}
\bibliography{custom}

\begin{thebibliography}{2}
\expandafter\ifx\csname natexlab\endcsname\relax\def\natexlab#1{#1}\fi

\bibitem[{Han et~al.(2018)Han, Zhu, Yu, Wang, Yao, Liu, and Sun}]{han2018fewrel}
Han, Xu and Zhu, Hao and Yu, Pengfei and Wang, Ziyun and Yao, Yuan and Liu, Zhiyuan and Sun, Maosong. 2018.
\newblock \emph{FewRel: A Large-Scale Supervised Few-Shot Relation Classification Dataset with State-of-the-Art Evaluation}.
\newblock ISLRN \href{https://www.islrn.org/resources/https://thunlp.github.io/1/fewrel1.html}{https://thunlp.github.io/1/fewrel1.html}.

\bibitem[{Zhang et~al.(2017)Zhang, Zhong, Chen, Angeli, and Manning}]{zhang2017position}
Zhang, Yuhao and Zhong, Victor and Chen, Danqi and Angeli, Gabor and Manning, Christopher D. 2017.
\newblock \emph{Position-aware Attention and Supervised Data Improve Slot Filling}.
\newblock ISLRN \href{https://www.islrn.org/resources/https://nlp.stanford.edu/projects/tacred/}{https://nlp.stanford.edu/projects/tacred/}.

\end{thebibliography}


\begin{thebibliography}{45}
\expandafter\ifx\csname natexlab\endcsname\relax\def\natexlab#1{#1}\fi

\bibitem[{Brown et~al.(2020)Brown, Mann, Ryder, Subbiah, Kaplan, Dhariwal, Neelakantan, Shyam, Sastry, Askell et~al.}]{brown2020language}
Tom Brown, Benjamin Mann, Nick Ryder, Melanie Subbiah, Jared~D Kaplan, Prafulla Dhariwal, Arvind Neelakantan, Pranav Shyam, Girish Sastry, Amanda Askell, et~al. 2020.
\newblock Language models are few-shot learners.
\newblock \emph{Advances in neural information processing systems}, 33:1877--1901.

\bibitem[{Chen et~al.(2022)Chen, Zhang, Xie, Deng, Yao, Tan, Huang, Si, and Chen}]{chen2022knowprompt}
Xiang Chen, Ningyu Zhang, Xin Xie, Shumin Deng, Yunzhi Yao, Chuanqi Tan, Fei Huang, Luo Si, and Huajun Chen. 2022.
\newblock Knowprompt: Knowledge-aware prompt-tuning with synergistic optimization for relation extraction.
\newblock In \emph{Proceedings of the ACM Web conference 2022}, pages 2778--2788.

\bibitem[{Chen et~al.(2023)Chen, Wu, and Shi}]{chen2023consistent}
Xiudi Chen, Hui Wu, and Xiaodong Shi. 2023.
\newblock Consistent prototype learning for few-shot continual relation extraction.
\newblock In \emph{Proceedings of the 61st Annual Meeting of the Association for Computational Linguistics (Volume 1: Long Papers)}, pages 7409--7422.

\bibitem[{Cui et~al.(2021)Cui, Yang, Yu, Hu, Cheng, Yi, and Xiao}]{cui2021refining}
Li~Cui, Deqing Yang, Jiaxin Yu, Chengwei Hu, Jiayang Cheng, Jingjie Yi, and Yanghua Xiao. 2021.
\newblock Refining sample embeddings with relation prototypes to enhance continual relation extraction.
\newblock In \emph{Proceedings of the 59th Annual Meeting of the Association for Computational Linguistics and the 11th International Joint Conference on Natural Language Processing (Volume 1: Long Papers)}, pages 232--243.

\bibitem[{Devlin et~al.(2019)Devlin, Chang, Lee, and Toutanova}]{devlin2019bert}
Jacob Devlin, Ming-Wei Chang, Kenton Lee, and Kristina Toutanova. 2019.
\newblock Bert: Pre-training of deep bidirectional transformers for language understanding.
\newblock In \emph{Proceedings of the 2019 Conference of the North American Chapter of the Association for Computational Linguistics: Human Language Technologies, Volume 1 (Long and Short Papers)}, pages 4171--4186.

\bibitem[{Dong et~al.(2022)Dong, Li, Dai, Zheng, Wu, Chang, Sun, Xu, and Sui}]{dong2022survey}
Qingxiu Dong, Lei Li, Damai Dai, Ce~Zheng, Zhiyong Wu, Baobao Chang, Xu~Sun, Jingjing Xu, and Zhifang Sui. 2022.
\newblock A survey for in-context learning.
\newblock \emph{arXiv preprint arXiv:2301.00234}.

\bibitem[{Fernando et~al.(2017)Fernando, Banarse, Blundell, Zwols, Ha, Rusu, Pritzel, and Wierstra}]{fernando2017pathnet}
Chrisantha Fernando, Dylan Banarse, Charles Blundell, Yori Zwols, David Ha, Andrei~A Rusu, Alexander Pritzel, and Daan Wierstra. 2017.
\newblock Pathnet: Evolution channels gradient descent in super neural networks.
\newblock \emph{arXiv preprint arXiv:1701.08734}.

\bibitem[{Gao et~al.(2021)Gao, Fisch, and Chen}]{gao2021making}
Tianyu Gao, Adam Fisch, and Danqi Chen. 2021.
\newblock Making pre-trained language models better few-shot learners.
\newblock In \emph{Joint Conference of the 59th Annual Meeting of the Association for Computational Linguistics and the 11th International Joint Conference on Natural Language Processing}, pages 3816--3830. Association for Computational Linguistics.

\bibitem[{Gu et~al.(2022)Gu, Han, Liu, and Huang}]{gu2022ppt}
Yuxian Gu, Xu~Han, Zhiyuan Liu, and Minlie Huang. 2022.
\newblock Ppt: Pre-trained prompt tuning for few-shot learning.
\newblock In \emph{Proceedings of the 60th Annual Meeting of the Association for Computational Linguistics (Volume 1: Long Papers)}, pages 8410--8423.

\bibitem[{Han et~al.(2020)Han, Dai, Gao, Lin, Liu, Li, Sun, and Zhou}]{han2020continual}
Xu~Han, Yi~Dai, Tianyu Gao, Yankai Lin, Zhiyuan Liu, Peng Li, Maosong Sun, and Jie Zhou. 2020.
\newblock Continual relation learning via episodic memory activation and reconsolidation.
\newblock In \emph{Proceedings of the 58th Annual Meeting of the Association for Computational Linguistics}, pages 6429--6440.

\bibitem[{Han et~al.(2022)Han, Zhao, Ding, Liu, and Sun}]{han2022ptr}
Xu~Han, Weilin Zhao, Ning Ding, Zhiyuan Liu, and Maosong Sun. 2022.
\newblock Ptr: Prompt tuning with rules for text classification.
\newblock \emph{AI Open}, 3:182--192.

\bibitem[{Hu et~al.(2022)Hu, Yang, Jin, Chen, and Xiao}]{hu2022improving}
Chengwei Hu, Deqing Yang, Haoliang Jin, Zhen Chen, and Yanghua Xiao. 2022.
\newblock Improving continual relation extraction through prototypical contrastive learning.
\newblock In \emph{Proceedings of the 29th International Conference on Computational Linguistics}, pages 1885--1895.

\bibitem[{Khosla et~al.(2020)Khosla, Teterwak, Wang, Sarna, Tian, Isola, Maschinot, Liu, and Krishnan}]{khosla2020supervised}
Prannay Khosla, Piotr Teterwak, Chen Wang, Aaron Sarna, Yonglong Tian, Phillip Isola, Aaron Maschinot, Ce~Liu, and Dilip Krishnan. 2020.
\newblock Supervised contrastive learning.
\newblock \emph{Advances in neural information processing systems}, 33:18661--18673.

\bibitem[{Kwiatkowski et~al.(2019)Kwiatkowski, Palomaki, Redfield, Collins, Parikh, Alberti, Epstein, Polosukhin, Devlin, Lee et~al.}]{kwiatkowski2019natural}
Tom Kwiatkowski, Jennimaria Palomaki, Olivia Redfield, Michael Collins, Ankur Parikh, Chris Alberti, Danielle Epstein, Illia Polosukhin, Jacob Devlin, Kenton Lee, et~al. 2019.
\newblock Natural questions: A benchmark for question answering research.
\newblock \emph{Transactions of the Association for Computational Linguistics}, 7:452--466.

\bibitem[{Lester et~al.(2021)Lester, Al-Rfou, and Constant}]{lester2021power}
Brian Lester, Rami Al-Rfou, and Noah Constant. 2021.
\newblock The power of scale for parameter-efficient prompt tuning.
\newblock In \emph{Proceedings of the 2021 Conference on Empirical Methods in Natural Language Processing}, pages 3045--3059.

\bibitem[{Li and Hoiem(2017)}]{li2017learning}
Zhizhong Li and Derek Hoiem. 2017.
\newblock Learning without forgetting.
\newblock \emph{IEEE transactions on pattern analysis and machine intelligence}, 40(12):2935--2947.

\bibitem[{Likas et~al.(2003)Likas, Vlassis, and Verbeek}]{likas2003global}
Aristidis Likas, Nikos Vlassis, and Jakob~J Verbeek. 2003.
\newblock The global k-means clustering algorithm.
\newblock \emph{Pattern recognition}, 36(2):451--461.

\bibitem[{Luo et~al.(2023)Luo, Yang, Meng, Li, Zhou, and Zhang}]{luo2023empirical}
Yun Luo, Zhen Yang, Fandong Meng, Yafu Li, Jie Zhou, and Yue Zhang. 2023.
\newblock An empirical study of catastrophic forgetting in large language models during continual fine-tuning.
\newblock \emph{arXiv preprint arXiv:2308.08747}.

\bibitem[{Mai et~al.(2021)Mai, Li, Kim, and Sanner}]{mai2021supervised}
Zheda Mai, Ruiwen Li, Hyunwoo Kim, and Scott Sanner. 2021.
\newblock Supervised contrastive replay: Revisiting the nearest class mean classifier in online class-incremental continual learning.
\newblock In \emph{Proceedings of the IEEE/CVF Conference on Computer Vision and Pattern Recognition}, pages 3589--3599.

\bibitem[{Mallya et~al.(2018)Mallya, Davis, and Lazebnik}]{mallya2018piggyback}
Arun Mallya, Dillon Davis, and Svetlana Lazebnik. 2018.
\newblock Piggyback: Adapting a single network to multiple tasks by learning to mask weights.
\newblock In \emph{Proceedings of the European conference on computer vision}, pages 67--82.

\bibitem[{Manning et~al.(2008)Manning, Raghavan, and Sch{\"u}tze}]{manning2008introduction}
Christopher~D Manning, Prabhakar Raghavan, and Hinrich Sch{\"u}tze. 2008.
\newblock Introduction to information retrieval.

\bibitem[{McCloskey and Cohen(1989)}]{mccloskey1989catastrophic}
Michael McCloskey and Neal~J Cohen. 1989.
\newblock Catastrophic interference in connectionist networks: The sequential learning problem.
\newblock \emph{Psychology of Learning and Motivation-Advances in Research and Theory}, 24(C):109--165.

\bibitem[{Oord et~al.(2018)Oord, Li, and Vinyals}]{oord2018representation}
Aaron van~den Oord, Yazhe Li, and Oriol Vinyals. 2018.
\newblock Representation learning with contrastive predictive coding.
\newblock \emph{arXiv preprint arXiv:1807.03748}.

\bibitem[{OpenAI(2022)}]{openai2022chatgpt}
OpenAI. 2022.
\newblock Chatgpt: Optimizing language models for dialogue.
\newblock In \emph{https://openai.com/blog}.

\bibitem[{OpenAI(2023{\natexlab{a}})}]{openai2023chat35}
OpenAI. 2023{\natexlab{a}}.
\newblock gpt-3.5-turbo.
\newblock In \emph{https://platform.openai.com/docs/models/gpt-3-5}.

\bibitem[{OpenAI(2023{\natexlab{b}})}]{openai2023gpt4}
OpenAI. 2023{\natexlab{b}}.
\newblock Gpt-4 technical report.
\newblock In \emph{https://arxiv.org/abs/2303.08774}.

\bibitem[{Paszke et~al.(2019)Paszke, Gross, Massa, Lerer, Bradbury, Chanan, Killeen, Lin, Gimelshein, Antiga et~al.}]{paszke2019pytorch}
Adam Paszke, Sam Gross, Francisco Massa, Adam Lerer, James Bradbury, Gregory Chanan, Trevor Killeen, Zeming Lin, Natalia Gimelshein, Luca Antiga, et~al. 2019.
\newblock Pytorch: An imperative style, high-performance deep learning library.
\newblock \emph{Advances in neural information processing systems}, 32.

\bibitem[{Peng et~al.(2020)Peng, Gao, Han, Lin, Li, Liu, Sun, and Zhou}]{peng2020learning}
Hao Peng, Tianyu Gao, Xu~Han, Yankai Lin, Peng Li, Zhiyuan Liu, Maosong Sun, and Jie Zhou. 2020.
\newblock Learning from context or names? an empirical study on neural relation extraction.
\newblock In \emph{Proceedings of the 2020 Conference on Empirical Methods in Natural Language Processing}, pages 3661--3672.

\bibitem[{Qin and Joty(2022)}]{qin2022continual}
Chengwei Qin and Shafiq Joty. 2022.
\newblock Continual few-shot relation learning via embedding space regularization and data augmentation.
\newblock In \emph{Proceedings of the 60th Annual Meeting of the Association for Computational Linguistics (Volume 1: Long Papers)}, pages 2776--2789.

\bibitem[{Rebuffi et~al.(2017)Rebuffi, Kolesnikov, Sperl, and Lampert}]{rebuffi2017icarl}
Sylvestre-Alvise Rebuffi, Alexander Kolesnikov, Georg Sperl, and Christoph~H Lampert. 2017.
\newblock icarl: Incremental classifier and representation learning.
\newblock In \emph{Proceedings of the IEEE conference on Computer Vision and Pattern Recognition}, pages 2001--2010.

\bibitem[{Ritter et~al.(2018)Ritter, Botev, and Barber}]{ritter2018online}
Hippolyt Ritter, Aleksandar Botev, and David Barber. 2018.
\newblock Online structured laplace approximations for overcoming catastrophic forgetting.
\newblock \emph{Advances in Neural Information Processing Systems}, 31.

\bibitem[{Schick and Sch{\"u}tze(2021)}]{schick2021exploiting}
Timo Schick and Hinrich Sch{\"u}tze. 2021.
\newblock Exploiting cloze-questions for few-shot text classification and natural language inference.
\newblock In \emph{Proceedings of the 16th Conference of the European Chapter of the Association for Computational Linguistics: Main Volume}, pages 255--269.

\bibitem[{Schroff et~al.(2015)Schroff, Kalenichenko, and Philbin}]{schroff2015facenet}
Florian Schroff, Dmitry Kalenichenko, and James Philbin. 2015.
\newblock Facenet: A unified embedding for face recognition and clustering.
\newblock In \emph{Proceedings of the IEEE conference on computer vision and pattern recognition}, pages 815--823.

\bibitem[{Shin et~al.(2017)Shin, Lee, Kim, and Kim}]{shin2017continual}
Hanul Shin, Jung~Kwon Lee, Jaehong Kim, and Jiwon Kim. 2017.
\newblock Continual learning with deep generative replay.
\newblock \emph{Advances in neural information processing systems}, 30.

\bibitem[{Sohn(2016)}]{sohn2016improved}
Kihyuk Sohn. 2016.
\newblock Improved deep metric learning with multi-class n-pair loss objective.
\newblock \emph{Advances in neural information processing systems}, 29.

\bibitem[{Song et~al.(2023)Song, Wang, Xiong, Zhu, Liu, Sui, and Li}]{song2023infocl}
Yifan Song, Peiyi Wang, Weimin Xiong, Dawei Zhu, Tianyu Liu, Zhifang Sui, and Sujian Li. 2023.
\newblock Infocl: Alleviating catastrophic forgetting in continual text classification from an information theoretic perspective.
\newblock In \emph{Findings of the Association for Computational Linguistics: EMNLP 2023}, pages 14557--14570.

\bibitem[{Van~der Maaten and Hinton(2008)}]{van2008visualizing}
Laurens Van~der Maaten and Geoffrey Hinton. 2008.
\newblock Visualizing data using t-sne.
\newblock \emph{Journal of machine learning research}, 9(11).

\bibitem[{Wang et~al.(2019)Wang, Xiong, Yu, Guo, Chang, and Wang}]{wang2019sentence}
Hong Wang, Wenhan Xiong, Mo~Yu, Xiaoxiao Guo, Shiyu Chang, and William~Yang Wang. 2019.
\newblock Sentence embedding alignment for lifelong relation extraction.
\newblock In \emph{Proceedings of the 2019 Conference of the North American Chapter of the Association for Computational Linguistics: Human Language Technologies, Volume 1 (Long and Short Papers)}, pages 796--806.

\bibitem[{Wang et~al.(2022)Wang, Song, Liu, Lin, Cao, Li, and Sui}]{wang2022learning}
Peiyi Wang, Yifan Song, Tianyu Liu, Binghuai Lin, Yunbo Cao, Sujian Li, and Zhifang Sui. 2022.
\newblock Learning robust representations for continual relation extraction via adversarial class augmentation.
\newblock In \emph{Proceedings of the 2022 Conference on Empirical Methods in Natural Language Processing}, pages 6264--6278.

\bibitem[{Wang et~al.(2023)Wang, Wang, and Hu}]{wang2023serial}
Xinyi Wang, Zitao Wang, and Wei Hu. 2023.
\newblock Serial contrastive knowledge distillation for continual few-shot relation extraction.
\newblock In \emph{Findings of the Association for Computational Linguistics}, pages 12693--12706.

\bibitem[{Wei et~al.(2022)Wei, Wang, Schuurmans, Bosma, Xia, Chi, Le, Zhou et~al.}]{wei2022chain}
Jason Wei, Xuezhi Wang, Dale Schuurmans, Maarten Bosma, Fei Xia, Ed~Chi, Quoc~V Le, Denny Zhou, et~al. 2022.
\newblock Chain-of-thought prompting elicits reasoning in large language models.
\newblock \emph{Advances in Neural Information Processing Systems}, 35:24824--24837.

\bibitem[{Wolf et~al.(2020)Wolf, Debut, Sanh, Chaumond, Delangue, Moi, Cistac, Rault, Louf, Funtowicz et~al.}]{wolf2020transformers}
Thomas Wolf, Lysandre Debut, Victor Sanh, Julien Chaumond, Clement Delangue, Anthony Moi, Pierric Cistac, Tim Rault, R{\'e}mi Louf, Morgan Funtowicz, et~al. 2020.
\newblock Transformers: State-of-the-art natural language processing.
\newblock In \emph{Proceedings of the 2020 conference on empirical methods in natural language processing: system demonstrations}, pages 38--45.

\bibitem[{Zhai et~al.(2023)Zhai, Tong, Li, Cai, Qu, Lee, and Ma}]{zhai2023investigating}
Yuexiang Zhai, Shengbang Tong, Xiao Li, Mu~Cai, Qing Qu, Yong~Jae Lee, and Yi~Ma. 2023.
\newblock Investigating the catastrophic forgetting in multimodal large language models.
\newblock \emph{arXiv preprint arXiv:2309.10313}.

\bibitem[{Zhang et~al.(2022)Zhang, Liang, Yang, Wang, and Xu}]{zhang2022prompt}
Han Zhang, Bin Liang, Min Yang, Hui Wang, and Ruifeng Xu. 2022.
\newblock Prompt-based prototypical framework for continual relation extraction.
\newblock \emph{IEEE/ACM Transactions on Audio, Speech, and Language Processing}, 30:2801--2813.

\bibitem[{Zhao et~al.(2022)Zhao, Xu, Yang, and Gao}]{zhao2022consistent}
Kang Zhao, Hua Xu, Jiangong Yang, and Kai Gao. 2022.
\newblock Consistent representation learning for continual relation extraction.
\newblock In \emph{Findings of the Association for Computational Linguistics}, pages 3402--3411.

\end{thebibliography}

\nocite{*}
\section{Language Resource References}
\bibliographystylelanguageresource{lrec-coling2024-natbib}
\bibliographylanguageresource{languageresource}

\appendix

\section{Experimental Details}
\subsection{Datasets} \label{data1}
We conduct experiments on two RE datasets: FewRel \citeplanguageresource{han2018fewrel} and TACRED \citeplanguageresource{zhang2017position}.

\paragraph{FewRel}
is a recently popular few-shot relation extraction dataset built from Wikidata.
It has high-quality manually annotated data, which contains 100 relations and 700 instances per relation. 
It is also a dataset with evenly distributed instances for each relation, without long-tail distribution issues.

\paragraph{TACRED}
is one of the most widely used relation extraction datasets with 106,264 examples and 41 relation types and \textit{no\_relation} if on defined relation if held.
It is built over newswire and web text from the corpus used in the yearly TAC Knowledge Base Population (TAC KBP) challenges \citeplanguageresource{zhang2017position}.
Unlike FewRel, TACRED has an unbalanced distribution of relations, which simulates real-world scenarios.
Based on the open relation assumption of CRE \cite{cui2021refining}, models continually learn certain relation types, we filter out \textit{no\_relation} type as previous works \cite{han2020continual, cui2021refining, qin2022continual, wang2023serial}.

\subsection{Baselines}
SCKD \cite{wang2023serial} also focuses on CFRE, but different from our work:
First, they focus on using knowledge distillation to preserve the prior knowledge, thus addressing catastrophic forgetting. Although it works, their approach requires extra computation and consumes more computing resources. While we concentrate more on how to better leverage PLMs to learn more generalized knowledge that can be easily adapted to old and new categories, thus activates catastrophic forgetting without extra computation cost.
Second, they use bidirectional data augmentation to generate pseudo samples, which essentially replace similar entities to generate new samples. Their augmented samples are unnatural and noisy and also lack diversity. We exploit the power of LLMs to boost smaller PLMs and propose a memory augmentation strategy that can generate more natural and diverse samples. 
Another work ConPL \cite{chen2023consistent} for CFRE, is not adopted for comparison due to its different task settings. 

\subsection{Implementation Details and Hyperparameters} \label{moreparams}
EMAR+ACA \cite{wang2022learning} and InfoCL \cite{song2023infocl} are designed for CRE, not CFRE. We re-implement their origin code in our few-shot setting with the default hyperparameters.
For task consistency, we re-implement their data processing using our data to keep the task sequence exactly the same as ours.
Note that, we set the temperature parameter in OpenAI API to 0, to ensure the stability of ChatGPT generation as much as possible, so the experimental results can be reproduced as much as possible.
However, there is no guarantee that ChatGPT generation will remain the same in the future.

We utilize 1 single NVIDIA Tesla P40 GPU with 24 GB memory on an Intel(R) Xeon(R) Gold 5118 CPU @2.30GHz to run all experiments.
For the hyperparameters search, we conduct a grid search to choose the appropriate values. 
Hyperparameters are illustrated in Table \ref{Hyperparameters}.

\begin{table}
    \centering
    \begin{tabular}{lc}
        \toprule
        Hyperparameters & Values \\
        \midrule
        seed & 100 \\
        batch\_size & 16 \\
        epoch\_currenTrain & 10 \\
        epoch\_memTrain & 10 \\
        learning\_rate & $1e-5$ \\
        optimizer & Adam\\
        \midrule
        BERT\_hidden\_size & 768 \\
        Encoder\_output\_size & 768 \\
        BERT\_input\_max\_length & 256 \\
        \midrule
        margin\_m & 0.3 \\
        normalization\_k & 0.5 \\
        temperature\_$\tau$ & 0.1 \\
        contrastive\_sample\_number & 500 \\
        \midrule
        soft\_prompt\_initialization & random \\
        soft\_prompt\_length & 3 \\
        soft\_prompt\_number & 4 \\
        \midrule
        ChatGPT\_temperature & 0 \\
        generated\_number\_FewRel & 2 \\
        generated\_number\_TACRED & 5 \\
        \bottomrule
    \end{tabular}
    \caption{Hyperparameters setting.}
    \label{Hyperparameters}
\end{table}

\begin{table*}
  \centering
  \small
  \scalebox{0.9}{
  \begin{tabular}{lcccccccc}
    \toprule
    \multicolumn{9}{c}{\textbf{FewRel} (\textit{10-way 10-shot})}\\
    \midrule
    \textbf{Method} & $\bm{\mathcal{T}^1}$ & $\bm{\mathcal{T}^2}$ & $\bm{\mathcal{T}^3}$ & $\bm{\mathcal{T}^4}$ & $\bm{\mathcal{T}^5}$ & $\bm{\mathcal{T}^6}$ & $\bm{\mathcal{T}^7}$ & $\bm{\mathcal{T}^8}$ \\ 
    \midrule
    Finetune & 95.43 & 45.67 & 28.26 & 21.69 & 17.94 & 13.83 & 12.24 & 10.70 \\
    Joint-train & \textbf{95.55} & \textbf{88.75} & \textbf{83.98} & \textbf{80.70} & \textbf{79.39} & \textbf{78.24} & \textbf{76.48} & \textbf{74.93} \\
    \midrule
    EMAR+ACA\textsuperscript{\ddag} \cite{wang2022learning} & 95.43 & 84.23 & 74.67 & 71.00 & 68.61 & 65.39 & 63.48 & 61.75 \\
    InfoCL\textsuperscript{\ddag} \cite{song2023infocl} & 95.08 & 83.34 & 76.03 & 72.05 & 70.09 & 66.97 & 65.09 & 63.02 \\
    RP-CRE\textsuperscript{\dag} \cite{cui2021refining} & 95.19 & 79.21 & 74.72 & 71.39 & 67.62 & 64.43 & 63.08 & 61.46 \\
    CRL\textsuperscript{\dag} \cite{zhao2022consistent} & 95.01 & 82.08 & 79.52 & 75.48 & 69.41 & 66.49 & 64.86 & 62.95 \\
    CRECL\textsuperscript{\dag} \cite{hu2022improving} & 95.63 & 83.81 & 78.06 & 71.28 & 68.32 & 66.76 & 64.95 & 63.01 \\
    ERDA\textsuperscript{\dag} \cite{qin2022continual} & 92.68 & 66.59 & 56.33 & 48.62 & 40.51 & 37.21 & 36.39 & 33.51 \\
    SCKD\textsuperscript{\ddag} \cite{wang2023serial}  & 95.43 & 86.51 & 79.72 & 76.01 & 73.69 & 70.45 & 68.22 & 66.58 \\
    \textbf{CPL (ours)} & \underline{95.54} & \underline{88.51} & \underline{81.92} & \underline{77.42} & \underline{74.71} & \underline{72.25} & \underline{69.63} & \underline{67.49} \\
    \midrule
    \midrule
    \multicolumn{9}{c}{\textbf{TACRED} (\textit{5-way 10-shot})}\\
    \midrule
    \textbf{Method} & $\bm{\mathcal{T}^1}$ & $\bm{\mathcal{T}^2}$ & $\bm{\mathcal{T}^3}$ & $\bm{\mathcal{T}^4}$ & $\bm{\mathcal{T}^5}$ & $\bm{\mathcal{T}^6}$ & $\bm{\mathcal{T}^7}$ & $\bm{\mathcal{T}^8}$ \\ 
    \midrule
    Finetune & 86.13 & 25.17 & 19.82 & 17.71 & 15.25 & 10.71 & 10.12 & 8.45 \\
    Joint-train & \textbf{88.37} & \textbf{83.79} & \textbf{76.75} & \textbf{72.15} & \textbf{70.21} & \textbf{66.60} & \textbf{64.29} & \textbf{61.42} \\
    \midrule
    EMAR+ACA\textsuperscript{\ddag} \cite{wang2022learning}  & 86.10 & 75.68 & 51.03 & 42.39 & 47.26 & 45.69 & 45.02 & 41.51 \\
    InfoCL\textsuperscript{\ddag} \cite{song2023infocl} & 85.42 & 74.64 & 62.49 & 57.80 & 54.65 & 49.81 & 46.24 & 44.25 \\
    RP-CRE\textsuperscript{\dag} \cite{cui2021refining} & 86.68 & 78.43 & 69.43 & 60.71 & 55.84 & 51.17 & 47.27 & 47.16 \\
    CRL\textsuperscript{\dag} \cite{zhao2022consistent} & 87.81 & 77.68 & 63.31 & 56.51 & 53.21 & 52.42 & 48.54 & 46.46 \\
    CRECL\textsuperscript{\dag} \cite{hu2022improving} & 83.88 & 73.45 & 59.24 & 53.51 & 49.27 & 47.41 & 45.15 & 44.33 \\
    ERDA\textsuperscript{\dag} \cite{qin2022continual} & 79.37 & 51.28 & 36.97 & 29.39 & 27.80 & 25.18 & 24.47 & 22.37 \\
    SCKD\textsuperscript{\ddag} \cite{wang2023serial} & \underline{88.28} & 80.77 & 72.41 & 65.57 & 64.90 & 58.60 & 55.65 & 53.42 \\
    \textbf{CPL (ours)} & 86.52 & \underline{81.78} & \underline{75.55} & \underline{68.04} & \underline{66.65} & \underline{61.72} & \underline{60.61} & \underline{58.57} \\
    \bottomrule
  \end{tabular}
  }
  \caption{Main results on FewRel and TACRED in 10-shot setting. \dag \, are reported in \cite{wang2023serial}, \ddag \, are we re-running the origin code. The \textbf{best results} are in bold, and the \underline{second-highest} is underlined.}
  \label{10shot}
\end{table*}

\section{More Experimental Results}
\subsection{Influence of Memory Size}
Memory size is an important factor for memory-based continual learning models.
Since CRE assumes that the model continues to learn new tasks, the number of tasks(samples) may be very large in the real-world scenario, and the memory size is limited, so it can not save more samples, but only save the key sample. Generally speaking, memory sampling and augmentation are employed to ensure that CFRE works practically in real-world scenarios.

\begin{figure}[htbp]
  \centering
  \includegraphics[width=0.45\textwidth]{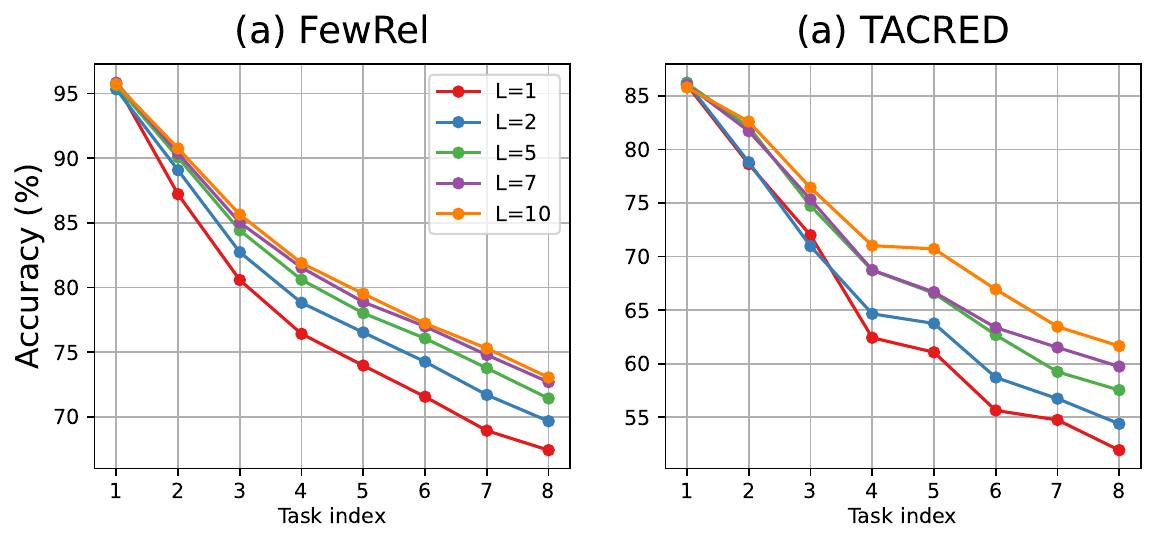}
  \caption{Results on memory size.}
  \label{memory}
\end{figure}

We set the memory size to 1 for a fair comparison with other baselines. We also did experiments to analyze the memory size, and all results are reported in the 10-shot setting without data generation. 
From figure \ref{memory}, with the increase in memory size, the performance of the models improves significantly, which proves the importance of memory size for continual learning. Notably, when the memory size increases to 10, that is, the memory contains all previous samples, and the performance is almost close to the Joint-train result.

\subsection{Results with 10-shot} \label{10shotresult}
Table \ref{10shot} shows the 10-shot results on FewRel and TACRED datasets.
From the results, our method still achieved the best performance in the 10-shot setting, which indicates that our method can alleviate catastrophic forgetting and overfitting even with more training samples.

\section{Generated Samples from ChatGPT} \label{gencase}
To generate more diverse examples, you can set the temperature parameter in OpenAI API to 1, to ensure that the samples generated are as diverse as possible.
But for the stability of the results, we set the temperature to 0 in our experiments.
The memory size is set to 1, so we only have one example as the demonstration in the input.
The prompt input is shown in Table \ref{case}, and we give some cases on FewRel dataset in Table \ref{fewrelcasegen}. 

From the results, we have some findings: 
(1) Our method is able to generate some high-quality samples, but the generated samples are still a little monotonous even though we've tried to make sure that the generated samples are as diverse as possible.
With relation \textit{``child''}, the words "mother", "father", "son" and "daughter" appear in almost all of the samples.
(2) Without example in the input, ChatGPT sometimes doesn't understand the relation well and generates some lexically similar samples, but actually false relations.

\begin{table*}
  \centering
  \begin{tabularx}{\textwidth}{p{0.15\textwidth} X}
    \toprule
        \textbf{Example} & \textbf{Cases} \\
    \midrule
        with & 
        Relation: child \newline
        Context: Michelle Obama is the \underline{mother} of Sasha and Malia Obama, both of whom are well-known for their activism and public speaking engagements. \newline
        Head Entity: Michelle Obama \newline
        Tail Entity: Sasha and Malia Obama \\
    \midrule
        with & 
        Relation: child \newline
        Context: Queen Elizabeth II is the \underline{mother} of Prince Charles, Prince Andrew, Princess Anne, and Prince Edward, all of whom hold royal titles and duties. \newline
        Head Entity: Queen Elizabeth II \newline
        Tail Entity: Prince Charles, Prince Andrew, Princess Anne, and Prince Edward \\    
    \midrule
        with & 
        Relation: child \newline
        Context: Renowned actor Will Smith is the \underline{father} of Jaden and Willow Smith, both of whom have followed in their father's footsteps and pursued careers in the entertainment industry. \newline
        Head Entity: Will Smith \newline
        Tail Entity: Jaden and Willow Smith \\           
    \midrule
        with & 
        Relation: child \newline
        Context: The famous singer Beyonce is the \underline{mother} of Blue Ivy, Sir, and Rumi Carter, all of whom have captured the hearts of fans with their appearances in music videos and public events. \newline
        Head Entity: Beyonce \newline
        Tail Entity: Blue Ivy, Sir, and Rumi Carter \\         
    \midrule
        with & 
        Relation: child \newline
        Context: Former President Barack Obama and First Lady Michelle Obama are the \underline{parents} of Malia and Sasha Obama, both of whom have grown up in the public eye and captivated audiences with their poise and intelligence. \newline
        Head Entity: Barack Obama and Michelle Obama \newline
        Tail Entity: Malia and Sasha Obama \\    
    \midrule
        without & 
        Relation: child \newline
        Context: Sarah watched proudly as her \underline{daughter}, Emily, rode her bike without training wheels for the first time. \newline
        Head Entity: Sarah \newline
        Tail Entity: Emily \\    
    \midrule
        without & 
        \textbf{Relation: child} \newline
        Context: John and Mary are \underline{siblings}. John is the eldest child in the family. \newline
        Head Entity: John \newline
        Tail Entity: Mary \\    
    \midrule
        without & 
        Relation: child \newline
        Context: The park was filled with families enjoying the sunny day, including Jane pushing her \underline{son}, Noah, on the swings. \newline
        Head Entity: Jane \newline
        Tail Entity: Noah \\    
    \midrule
        without & 
        Relation: child \newline
        Context: The school auditorium was buzzing with excitement as Maria prepared to watch her \underline{daughter}, Isabella, perform in the school play. \newline
        Head Entity: Maria \newline
        Tail Entity: Isabella \\    
    \midrule
        without & 
        Relation: child \newline
        Context: Jake beamed with pride as he watched his \underline{son}, Liam, take his first steps towards him. \newline
        Head Entity: Jake \newline
        Tail Entity: Liam \\    
    \bottomrule
  \end{tabularx}
  \caption{Generated cases on FewRel. \textbf{Errors} are highlighted in bold and \underline{keywords} for judging relations are underlined.}
  \label{fewrelcasegen}
\end{table*}

\end{document}